%% file: 0.main.tex
\documentclass[conference]{IEEEtran}
\IEEEoverridecommandlockouts
\usepackage{enumitem}
\usepackage{algorithm}
\usepackage{etoolbox}
\AtBeginEnvironment{algorithm}{\small}
\AtBeginEnvironment{algorithmic}{\small}

\usepackage[noend]{algpseudocode}
\usepackage{makecell}
\usepackage{multirow}
\usepackage{color, colortbl}
\usepackage{comment}
\usepackage{pifont}
\usepackage{graphicx}
\usepackage{booktabs}
\usepackage{amsmath}
\usepackage{amssymb}
\usepackage{amsfonts}
\usepackage{xcolor}
\usepackage{caption}
\usepackage{subcaption}
\usepackage{url}
\usepackage{enumitem}
\definecolor{lightgray}{gray}{0.9}

\AtEndPreamble{
    \usepackage[capitalize]{cleveref}
    \crefname{section}{Sec.}{Secs.}
    \Crefname{section}{Section}{Sections}
    \Crefname{table}{Table}{Tables}
    \crefname{table}{Tab.}{Tabs.}
}

\def\BibTeX{{\rm B\kern-.05em{\sc i\kern-.025em b}\kern-.08em
    T\kern-.1667em\lower.7ex\hbox{E}\kern-.125emX}}
\begin{document}

\let\titleold\title
\renewcommand{\title}[1]{\titleold{#1}\newcommand{\thetitle}{#1}}
\def\maketitlesupplementary
   {
   \newpage
       \twocolumn[
        \centering
        \Large
        \textbf{\thetitle}\\
        \vspace{0.5em}Supplementary Material \\
        \vspace{1.0em}
       ] 
   }

\title{Knowledge Priors for Identity-Disentangled Open-Set Privacy-Preserving Video FER}

\author{Feng Xu,
        Xun Li,
        Lars Petersson, 
        Yulei Sui, 
        David Ahmedt-Aristizabal, 
        Dadong Wang
\thanks{F. Xu is with the School of CSE, UNSW Sydney, and also with CSIRO's Data61, Australia (e-mail: feng.xu2@unsw.edu.au)}
\thanks{X. Li, L. Petersson, D. Ahmedt-Aristizabal and D. Wang are with CSIRO's Data61, Australia}
\thanks{Y. Sui is with the School of CSE, UNSW Sydney, Australia}}

\maketitle

\begin{abstract}
Facial expression recognition (FER) relies on facial data that inherently expose identity and thus raise significant privacy concerns. Current privacy-preserving methods typically fail in realistic open-set video settings where identities are unknown, and identity labels are unavailable. We propose a two-stage framework for video-based privacy-preserving FER in challenging open-set settings that requires no identity labels at any stage. To decouple privacy and utility, we first train an identity-suppression network using intra- and inter-video knowledge priors derived from real-world videos without identity labels. This network anonymizes identity while preserving expressive cues. A subsequent denoising module restores expression-related information and helps recover FER performance. Furthermore, we introduce a falsification-based validation method that uses recognition priors to rigorously evaluate privacy robustness without requiring annotated identity labels. Experiments on three video datasets demonstrate that our method effectively protects privacy while maintaining FER accuracy comparable to identity-supervised baselines.

\end{abstract}


\input{1.intro}

\input{2.related}
\input{3.method}

\input{4.exp}

\input{5.concl}

\bibliographystyle{bibstyle}
\bibliography{0.ref}

\input{6.supp}

\end{document}

%% file: 1.intro.tex
\section{Introduction}\label{intro}
 
Facial expression recognition (FER) in videos is essential for understanding human affect and behavior and supports a range of applications.
However, FER models rely on visual data that inherently reveals personal identity, raising legal, ethical, and regulatory concerns~\cite{voigt2017eu}. These risks hinder deployment of FER systems, especially in sensitive environments.
To address these concerns, many privacy-preserving methods have been proposed~\cite{wu2020privacy,dave2022spact,fioresi2023ted,xu2024facial}, designed to conceal identity while maintaining utility for downstream tasks.
Existing approaches can be grouped into image-based and video-based methods, and are often further categorized into two paradigms~\cite{osorio2021stable}: 
\textit{closed-set}, which assumes access to known identity labels during training~\cite{wu2020privacy, xu2024facial}, and 
\textit{open-set}, where such labels are unavailable and identities may be unseen during training or deployment~\cite{dave2022spact, fioresi2023ted}.

Despite progress, several limitations persist. 
Most work still focuses on still images~\cite{chen2018vgan}, even though real-world applications are video-based and require temporal modeling.
Existing video-based methods~\cite{xu2024facial} often rely on identity labels, which is impractical in open-set scenarios where such labels are unavailable. 
Adversarial approaches~\cite{wu2020privacy, dave2022spact} jointly optimize privacy and utility but often entangle identity and expression features, degrading both recognition accuracy and anonymization quality.
Evaluation is also challenging: existing evaluation protocols~\cite{xu2024facial, zhang2024validating} depend on identity supervision and do not generalize well to label-free open-set settings, where models must prevent identity classification and resist matching attacks against the original input.

\begin{figure}[!t]
    \centering
    \includegraphics[width=0.95\linewidth]{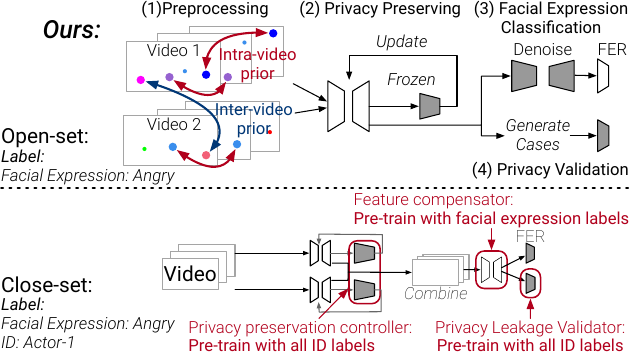}
    \caption{
     Our framework relies solely intra- and inter-video knowledge priors extracted from videos based on video characteristics to learn privacy preservation transformations without requiring identity labels. It integrates a denoising step to recover FER performance after anonymization and introduces a falsification-based validation scheme for privacy assessment.
    }
    \label{intro_fig}
\end{figure}

To address these challenges, we propose a privacy-preserving framework for video-based FER in open-set settings, as illustrated in~\cref{intro_fig}, along with a novel evaluation protocol for assessing privacy robustness.
Our framework includes four key components:
\textit{1. knowledge prior extraction and dataset generation:} intra- and inter-video priors are derived from unlabeled videos to guide training;
\textit{2. identity-disentangled privacy preservation:} a prior-driven identity suppression network suppresses identity features while retaining expression-relevant information;
\textit{3. denoising and FER training:} a denoising module mitigates expression distortion introduced by anonymization;
\textit{4. human annotation-free validation:} a falsification-based scheme that evaluates privacy protection without identity labels.
Our method is compared with prior works in~\cref{pp_compare}, where ``Act. Recog.'' and ``N/A'' denote action recognition and not applicable, respectively.
In summary, our main contributions are:

\begin{table}[!b]
    \centering
    \caption{Comparison of our approach with existing privacy-preserving approaches with their associated utility tasks. 
    }
    \resizebox{0.99\linewidth}{!}{
    \begin{tabular}{c|c|c|c|c|c|c}
    \toprule
      \makecell{Privacy \\Preservation}& \makecell{Identity \\ Label} & Denoise & Utility Task & \makecell{Privacy\\ Validation} & \makecell{Validation \\ Require Label} & Scenario \\ \midrule
      \cite{wu2020privacy}  & Yes& No & Act. Recog. &  Yes & Yes & Closed\\
      \cite{dave2022spact}\cite{fioresi2023ted}  & No  & No & Act. Recog. & Yes & Yes & Open\\
      Blurring and \cite{xu2022mobilefaceswap} & No & No& N/A& No & N/A& Open\\
      \cite{xu2024facial}  & Yes& Yes & FER & Yes  & Yes  & Closed\\ \midrule
      Ours & No& Yes & FER & Yes  & No & Open\\      
       \bottomrule
    \end{tabular}
    }
    
    \label{pp_compare}
\end{table}

\begin{figure*}[!h]
    \centering
    \includegraphics[width=0.85\linewidth]{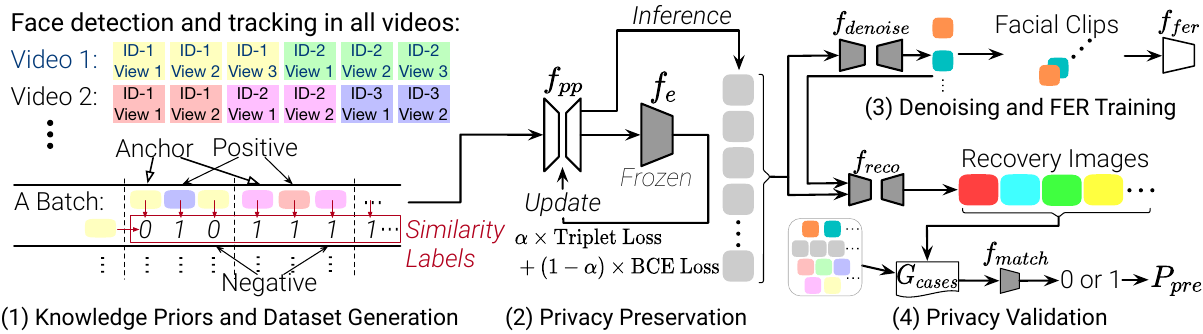}
    \caption{ 
    Overview of our privacy-preserving video-based FER framework. 
    (1) Face detection and tracking extract cropped face sequences and assign pseudo-identity labels. 
    (2) These labeled faces train the privacy-preserving reconstruction network ($f_{pp}$) using triplet and binary cross-entropy (BCE) losses, guided by the frozen identity extractor ($f_e$). 
    (3) $f_{pp}$ generates anonymized faces, which are enhanced by a denoising network $f_{denoise}$ to support FER training via ($f_{fer}$). 
    (4) Privacy robustness is evaluated using a falsification-based approach without identity labels, comparing original, anonymized, denoised, and reconstructed outputs via a recovery model ($f_{reco}$), with a privacy preservation ratio $P_{pre}$ computed for quantitative comparison.
    }
    \label{fig:framework}
\end{figure*}

\begin{itemize}
    \item A privacy-preserving video FER framework that leverages intra- and inter-video characteristics to learn identity-invariant representations without requiring identity labels.
    \item A knowledge-prior–based training strategy that extends privacy-preserving FER to open-set scenarios.
    \item A post-anonymization denoising module that enhances video-based FER accuracy.
    \item A falsification-based privacy evaluation protocol that operates without identity labels and supports open-set validation.
\end{itemize}

%% file: 2.related.tex
\section{Related Work}\label{rw}

\noindent\textbf{Privacy-preserved facial expression recognition.}
Recent work on balancing privacy preservation and FER has focused mainly on images. Image-level approaches~\cite{chen2018vgan} synthesize privacy-preserving facial images, while feature-level methods~\cite{leibl2023identifying} manipulate identity features to retain expression cues. 
FaceMotionPreserve~\cite{zhu2024facemotionpreserve} uses generative models to alter identity and enhance emotion signals. These methods, however, do not naturally extend to video sequences, where temporal coherence is important. In contrast, the only existing video-based approach~\cite{xu2024facial} separates frequency components for anonymization but requires identity labels.

\noindent\textbf{Privacy-preserved soft biometrics.}
Wu et al.~\cite{wu2020privacy} introduce an adversarial training framework that jointly optimizes privacy and utility using explicit privacy attribute labels. SPAct~\cite{dave2022spact} extends \cite{wu2020privacy} by removing the need for such labels through self-supervision, and Ted-SPAD~\cite{fioresi2023ted} extends SPAct to anomaly detection with a temporally-aware triplet loss. While these methods balance privacy and utility, adversarial training has been shown to be suboptimal for facial privacy tasks~\cite{xu2024facial}.

\noindent\textbf{Privacy preservation validation.}
$\mathrm{Map^2V}$~\cite{zhang2024validating} uses image priors and gradient-based estimation to assess privacy preservation under minimal assumptions, targeting adversarial attacks in a black-box setting. In~\cite{xu2024facial}, identity classifiers are trained on anonymized frames, and classification accuracy is used as a proxy for privacy leakage. While best suited to closed-set identity-supervised scenarios, this protocol remains a standard benchmark for comparing privacy-preserving methods.

%% file: 3.method.tex
\section{Method}
To address existing limitations, our framework 
(1) protects facial identity privacy in videos under open-set conditions without identity annotations, and 
(2) enables FER models to learn effectively from anonymized clips. 
Our framework also includes a validation mechanism to assess privacy preservation without relying on identity labels.

\subsection{Preliminary: Characteristics of Videos} \label{preliminary}

Real-world videos exhibit several key characteristics that guide the design of our privacy-preserving FER framework, as illustrated in ~\cref{fig:framework}. 
Each individual appears at most once per frame, and repeated appearances across unrelated videos are rare. A frame $F_{i,j}$ may contain multiple faces by $\mathbb{I}_{V_i}=\{I{i,j}^{1},..., I_{i,j}^{k},...\}$, where $i$ and $j$ index the video and frame, respectively, and a video $V_i={F_{i,j}}$ forms a sequence of frames, with $k$ indexing detected faces. 
People cannot ``teleport'': a face present in one frame persists across consecutive frames of the same video. Privacy protection is applied only to detected faces, while frames without detections are considered identity-free. Typical datasets (e.g., healthcare, HCI) do not contain hidden facial features requiring specialized detection. Finally, we assume an open-set setting~\cite{osorio2021stable}, where FER labels are available, but identity annotations are not.

\subsection{Knowledge Priors and Dataset Generation}

In the first stage, inter- and intra-video knowledge priors defines data relationships that guide the training of the privacy preservation encoder-decoder network $f_{pp}$. 
A face detection network is applied to all frames across all videos, and detected faces are then cropped and aligned.
Identity embeddings are then extracted and compared against a similarity threshold to enable robust face tracking and pseudo-label generation.
For each video, each unique tracked face is assigned a new tracking ID (pseudo label) that remains consistent throughout the video.
Each tracked face is thus associated with its video number, frame index, tracking ID, and bounding box coordinates.

\begin{figure}[!t]
    \centering
    \includegraphics[width=0.99\linewidth]{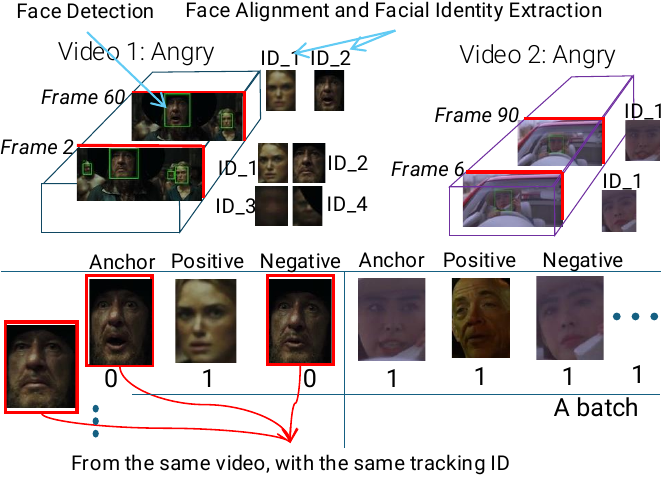}
    \caption{
    Illustration of inter- and intra-video knowledge priors for training $f_{pp}$. Each video undergoes face detection, alignment, and tracking to obtain consistent tracking IDs and pseudo-labels. Expression labels (e.g., ``Angry'') group semantically similar samples across videos. Training batches contain triplets (anchor, positive, negative) plus an extra face sharing the anchor’s ID to enforce identity disentanglement. A binary label (0 or 1) signals identity consistency, enabling identity-irrelevant learning without ground-truth ID labels.
    }
    \label{preprocessing_example}
\end{figure}

Second, these pseudo-labels and expression labels are used to construct batches for training $f_{pp}$, aiming to preserve expression features while suppressing identity. To achieve this without privacy attribute labels, i) anchor-negative pairs are selected from faces with the same tracking ID (same identity), ii) a randomly chosen face with the same expression label but from a different video serves as the positive, and iii) each batch contains several such triplets. Additionally, one more face sharing the tracking ID of the anchor of the first pair is included at the start of the batch. In~\cref{preprocessing_example}, we illustrate such a batch. The training on this dataset can incorporate two objective functions: a triplet loss and a binary cross-entropy loss (BCE). The triplet loss enforces expression feature similarity learning and suppress identify feature learning among faces sharing the same tracking ID. The BCE distinguishes identity similarity from dissimilarity: the extra face is labeled 0 with the anchor and negative (identity mismatch sharing the same tracking ID), and 1 with all remaining samples (identity dissimilarity under the constraint of expression similarity).

Our method offers two key advantages. First, unlike clustering-based methods that perform global clustering over all facial embeddings, being highly sensitive to large intra-identity variations in the wild, our approach derives pseudo-labels from tracking-based temporal consistency with strict similarity thresholds, avoiding unreliable cross-video comparisons and enabling more robust identity assignments. Secondly, while high similarity thresholds can effectively avoid identity merging, they may introduce identity switches. However, since each triplet in a batch is sampled from a different video, such switches do not affect dataset construction.

\subsection{Privacy Preservation}
We extend the closed-set training paradigm of~\cite{xu2024facial} to the open-set setting by decoupling privacy and utility objectives. As noted in~\cref{rw}, end-to-end adversarial learning struggles to jointly optimize privacy preservation and FER performance. In contrast, we adopt a two-stage strategy: we first train the privacy-preserving model using tailored batches and objectives to retain FER-relevant features, and then train the utility (FER) model separately on the anonymized data. Our method preserves privacy by reconstructing facial images using an encoder-decoder network, $f_{pp}$, as shown in~\cref{fig:framework}(2). The reconstructed faces are passed through a frozen identity extractor $f_e$, and we optimized $f_{pp}$ with triplet loss for anonymization and BCE loss to further enforce anonymization while preserving FER features. As detailed in \cref{pp_train}, training of the privacy model $f_{pp}$ uses the pre-processed data. We first pre-train $f_{pp}$ on cropped face images with an L1 reconstruction loss following~\cite{dave2022spact}. The identity extractor $f_e$ is adapted from VGG-Face by removing its classification head and initializing it with LFW-pretrained weights. During $f_{pp}$ training, $f_e$ remains frozen and provides identity embeddings for each batch. After training, $f_{pp}$ is applied to all datasets to produce privacy-preserved facial video clips.

\begin{algorithm} [!t]
\footnotesize
    \caption{Privacy Preservation Training}\label{pp_train}
\begin{algorithmic}[1]
    \Statex \textbf{Init:} Load Pre-trained Parameters for $f_{pp}$, $f_{e}$; an all-ones vector $labels$; Triplet Loss $L_{tri}$ and Binary Cross Entropy Loss $L_{bce}$;
    \Statex \textbf{Input:} $loader$ data from previous step. 
    \State Frozen parameters of $f_{e}$ (requires\_grad = False)
    \State Replace the values of index 0 and 2: $labels[[0,2]]=0$
    \For{$i$ in $loader$}
        \State \hspace*{-2mm}Privacy-preserved batch $i^r$ := $f_{pp} (i)$;
        \State \hspace*{-2mm}Extracted identity feature vector $fea$ := $f_{e} (i^r)$;
        \State \hspace*{-2mm}Calculate the cosine similarity within a batch $cos\_sim$, of $fea[1:]$ and $fea[0]$;
        \State \hspace*{-2mm}$l_{bce}$:=$L_{bce}(cos\_sim, \  labels)$;
        \State \hspace*{-2mm}$l_{tri}$:=$L_{tri}(fea[1::3],\ fea[2::3],\ fea[3::3])$
        \State \hspace*{-2mm}The total loss $l$:=$\alpha \times l_{tri} + (1-\alpha)\times l_{bce}$;
        \State \hspace*{-2mm}Backward and update $f_{pp}$ parameters;        
    
    \EndFor
    \Statex \textbf{Output:} $f_{pp}$
\end{algorithmic}

\end{algorithm}

\subsection{Denoising and Video-based FER}

Since both identity and expression features originate from the face, suppressing identity may inadvertently distort expression cues. To mitigate this, we introduce a denoising module, $f_{denoise}$, trained to restore expression features from privacy-preserved videos. The module is supervised using expression labels on $f_{pp}$-processed data from a separate in-the-wild dataset, and its outputs are passed through a fixed, pre-trained FER classifier; only $f_{denoise}$ is updated. Training details are provided in~\cref{denoise_train}. After training, $f_{denoise}$ is applied at inference to enhance privacy-preserved inputs, and the FER model is then trained with cross-entropy loss.

\begin{algorithm} [!t]
    \caption{Train $f_{denoise}$}\label{denoise_train}
\begin{algorithmic}[1]
    \Statex \textbf{Init:} $f_{denoise}$, $f_{exp}$ and $f_{pp}$;
    \Statex \textbf{Input:} Data from RAF-DB~\cite{li2017reliable} including data, $D_{raf-db}$ and facial expression label, $E_{raf-db}$;
    \State $f_{pp}$ infers $D_{raf-db}$, gets privacy-preserved $D_{raf-db}'$;
    \State Merges $D_{raf-db}'$ and $E_{raf-db}$ as $D_{raf-db}'$;
    \State Frozen parameters of $f_{exp}$ (requires\_grad = False);
    \For{each epoch}
        \For{each batch $d_{raf-db}'$ and $e_{raf-db}$ of $D_{raf-db}'$}
        \State \hspace*{-2mm}\textit{\footnotesize \# denoise image:}
        \State \hspace*{-2mm}$d_{raf-db}''$:=$f_{denoise}(d_{raf-db}')$;
        \State \hspace*{-2mm}\textit{\footnotesize \# FER classification:}
        \State \hspace*{-2mm}$e_{raf-db}''$:=$f_{exp}(d_{raf-db}'')$;
        \State \hspace*{-2mm}Cross Entropy Loss($e_{raf-db}''$, $e_{raf-db}$)
        \State \hspace*{-2mm}Update $f_{denoise}$ parameters
        \EndFor
    \EndFor
     \textbf{Output:} $f_{denoise}$
\end{algorithmic}
\end{algorithm}

\begin{table}[!t]
    \centering
    \caption{Test case generation rules. GT stands for ground truth. 0 means that $f_{match}$ needs to output the pair have different identities, otherwise is 1.}
    \resizebox{\linewidth}{!}{
    \begin{tabular}{c|l}
        \toprule
         GT & \multicolumn{1}{c}{Rules}  \\ \midrule
        \textit{0} & \makecell[l]{\textbf{1}. Two different images from the same  \textbf{original} \textbf{video} with different unique ID; \\ \textbf{2}. Two images from two different \textbf{original videos};}\\ \hline
         \textit{1} &\makecell[l]{\textbf{3}. Two images from the same \textbf{original video}  with the same unique ID;} \\ \hline
         0 &\makecell[l]{\textbf{4}. One image from original video and the other from\\ the same video but \textbf{privacy-preserved}, with the same unique ID;\\
         \textbf{5}. One image from original video and the other from the \\same video but \textbf{denoised privacy-preserved}, with the same unique ID;\\
         \textbf{6}. One image from original video and the other from the \\same video but \textbf{recovered privacy-preserved}, with the same unique ID;\\
         \textbf{7}. One image from original video and the other from the \\same video but \textbf{recovered denoised privacy-preserved}, with the same unique ID\\
         } \\ 
         \bottomrule
    \end{tabular}
    }
    
    \label{strategy}
\end{table}

\subsection{Validation of Privacy Preservation}
Most existing methods~\cite{xu2024facial, dave2022spact, wu2020privacy} rely on identity labels and thus are unsuitable for open-set scenarios without identity ground truth. To assess privacy-preserving approaches in open-set scenarios, we introduce a human annotation-free evaluation protocol that uses knowledge priors and a falsification-based approach, and also assesses robustness against recovery attacks. The design follows the characteristics in~\cref{preliminary}. Our validation consists of two components: (1) A rule-based case generator $G_{cases}$ that constructs identity-comparison pairs, and (2) a binary classifier $f_{match}$ that predicts whether two images originate from the same identity (1) or not (0).

To simulate a white-box adversary, we use the original dataset $D_{org}$, privacy-preserved data $D_{pp}$, and denoised outputs $D_{dpp}$. Two recovery models, $f_{reco}$, are trained to invert the privacy transformation using paired samples: original versus privacy-preserved faces sharing the same tracking ID, and original versus denoised privacy-preserved faces. Both models are trained with SSIM loss. After training, inference on $D_{pp}$ and $D_{dpp}$ yields recovered datasets $D_{pp,r}$ and $D_{dpp,r}$, enabling quantitative privacy evaluation.

The rule-based generator $G_{cases}$ constructs test cases from $D_{org}$, $D_{pp}$, $D_{dpp}$, $D_{pp,r}$ and $D_{dpp,r}$, following the rules in~\cref{strategy}. Rules 1-3 assess $f_{match}$ by evaluating its ability to classify facial identity, using binary ground-truth labels from same video pairs, with higher classification accuracy indicating better performance. Rules 4-7 apply a falsification strategy to privacy preservation: although paired images share the same unique ID, they should be dissimilar after anonymization, so the ground-truth labels are inverted. Pairs from different unique IDs are excluded because a prediction of 1 unambiguously indicates privacy leakage, whereas a prediction of 0 is inconclusive, as it may result from successful anonymization or from naturally dissimilar original identities.

The binary classifier $f_{match}$ is applied to all pairs generated by $G_{cases}$.
By comparing predictions with ground truth labels, we compute the privacy preservation ratio $P_{pre}$, defined as the proportion of correctly classified cases under Rules 4–7. A higher $P_{pre}$ indicates stronger privacy preservation, as it reflects a reduced ability of the model to correctly infer identities after anonymization. This metric supports direct comparison across privacy-preserving methods.

%% file: 4.exp.tex
\section{Experiments}\label{exp}

\subsection{Experimental settings}
\noindent\textbf{Datasets.} 
The proposed framework targets open-set, in-the-wild datasets without human-annotated identity labels. We conduct experiments on the DFEW dataset~\cite{jiang2020dfew} and additionally evaluate on two closed-set datasets with identity annotations, CREMA-D~\cite{cao2014crema} and RAVDESS~\cite{livingstone2012ryerson}. The denoising model is pre-trained on RAF-DB~\cite{li2017reliable}. Detailed dataset descriptions are provided in Supplementary Material~\cref{detailed_dataset}.

\noindent\textbf{Implementation.} \label{Implementation}
Face detection and alignment are performed with  RetinaFace~\cite{deng2019retinaface}, and identity embeddings are extracted with ArcFace~\cite{deng2019arcface} using a 0.7 cosine similarity threshold.
From DFEW, we extract 25,969 faces across 16,372 videos to generate knowledge priors, and privacy-preserving training uses a batch size of 1,024 (341 triplets), with alternatives analyzed in Supplementary Material ~\cref{sec:num_triplet}.
Privacy preservation employs U-Net as $f_{pp}$ and ArcFace~\cite{deng2019arcface} as $f_{e}$~\cref{pp_train}, with $\alpha=0.01$ over 400 epochs. $f_{pp}$ produces privacy-preserved cropped faces for validation and 11,697 facial video clips for denoising and FER training. 
The denoising model $f_{denoise}$ is pre-trained following~\cref{denoise_train} and applied to clips prior to FER training.

FER experiments follow the DFEW “set\_1” split~\cite{jiang2020dfew}, while CREMA-D and RAVDESS (7,442 and 4,904 faces) are privacy-preserved using the DFEW-trained $f_{pp}$, with 30\% of samples reserved for testing. FER backbones are two video understanding networks: R(2+1)D~\cite{tran2018closer} and I3D~\cite{carreira2017quo}, pre-trained on Kinetics-400~\cite{kay2017kinetics}. For privacy validation, $f_{reco}$ is a U-Net trained on $\langle i_{org}, i_{pp} \rangle$ and $\langle i_{org}, i_{dpp} \rangle$ pairs, and $f_{match}$ uses pre-trained ArcFace embeddings. The random seed is fixed at 42. Further details are in Supplementary Material~\cref{detailed_implementation}.

\subsection{Baselines}
We evaluate four privacy-preserving baselines, namely (1) \textit{blurring} (``GB''); (2) an \textit{adversarial privacy-preserved approach}~\cite{dave2022spact} (``Adver.''); (3) \textit{face swapping} via MobileFaceSwap~\cite{xu2022mobilefaceswap} (``Face S.''); and (4) a \textit{controlled high- and low-frequency approach}~\cite{xu2024facial}, (``Contr-HL''). Detailed descriptions of these baselines are provided in Supplementary Material~\cref{dba}. The first three, taken from Rows 2–3 in \cref{pp_compare}, operate in open-set settings; for these methods, we directly apply the corresponding obfuscation and then train the FER models without integrating our privacy-preserving module. The final baseline is implemented as described in its original work. All baselines except~\cite{xu2024facial}, which already includes feature compensation, use our denoising and recovery-attack modules during evaluation.

\subsection{Results and Evaluation}
\begin{table}[!t]
    \centering
    \caption{R(2+1)D and I3D FER accuracy (\%) on the testing parts of datasets \textbf{without privacy preservation} in regular font and \textbf{post privacy preservation and denoising} in bold and italics font.}
    \resizebox{\linewidth}{!}{
    \begin{tabular}{c|c|c|c|c|c|c|c|c|c}
    \toprule
       Dataset  & Model & Hap & Sad& Neu & Ang & Sup & Dis & Fea & Cal \\ \midrule
        \multirow{4}{*}{DFEW} & R(2+1)D & 79.96 & 40.11 & 56.93 & 50.11 & 49.32 & 3.45 & 22.10 & N/A  \\ 
        & I3D & 77.91 & 46.17 & 55.99 & 46.21 & 46.26 & 3.45 & 20.99 & N/A  \\ 
         & \textit{\textbf{R(2+1)D}} & \textit{\textbf{\textit{73.21}}} & \textit{\textbf{35.36}}  &  \textit{\textbf{\textit{47.57}}}  & \textit{\textbf{39.77}} & \textit{\textbf{38.78} }&\textit{ \textbf{3.45} }& \textit{\textbf{17.68}} & \textit{\textbf{N/A}}  \\ 
        & \textbf{\textit{I3D}} & \textit{\textbf{68.71}} & \textit{\textbf{39.31}} & \textit{\textbf{48.50}} & \textit{\textbf{40.23} }& \textit{\textbf{39.80}} & \textit{\textbf{3.45}} & \textit{\textbf{16.57}} & \textbf{\textit{N/A}}  \\ \midrule
        
        \multirow{4}{*}{CREMA-D}& R(2+1)D & 98.34 & 80.77 & 84.89 & 91.69 & N/A & 98.89 & 80.30 & N/A  \\ 
        & I3D & 99.17 & 82.21 & 87.31 & 90.03 & N/A & 93.35 & 77.50 &  N/A \\ 
        & \textbf{\textit{R(2+1)D}} & \textit{\textbf{91.44} }& \textit{\textbf{73.32}} & \textit{\textbf{76.13} }& \textit{\textbf{81.72}} & \textit{\textbf{N/A}} & \textit{\textbf{89.47}} & \textit{\textbf{73.32} }& \textit{\textbf{N/A}}  \\ 
        & \textbf{\textit{I3D}} & \textit{\textbf{93.09} }& \textit{\textbf{77.40}} & \textit{\textbf{84.05}} & \textit{\textbf{82.27}} & \textit{\textbf{N/A}} & \textit{\textbf{85.87}} & \textit{\textbf{70.00}} &  \textit{\textbf{N/A}} \\ \midrule
        
        \multirow{4}{*}{RAVDESS}& R(2+1)D & 88.47 & 70.24 & 89.09 & 89.88 & 89.75 & 91.34 & 82.31 & 90.40  \\ 
        & I3D & 93.56 & 82.35 & 82.18 & 92.26 & 88.93 & 88.09 & 88.10 & 86.36  \\
        
        & \textit{\textbf{R(2+1)D}} & \textit{\textbf{81.36}} & \textit{\textbf{67.82}} & \textit{\textbf{81.09}} & \textit{\textbf{83.33}} & \textit{\textbf{81.15}} & \textit{\textbf{84.12}} & \textit{\textbf{72.45}} & \textit{\textbf{82.32}}  \\
        & \textbf{\textit{I3D}} &\textit{ \textbf{86.44}} & \textit{\textbf{76.12}} & \textit{\textbf{78.55}} & \textit{\textbf{84.52}} & \textit{\textbf{79.92}} & \textit{\textbf{79.06}} & \textit{\textbf{80.95}} & \textit{\textbf{77.27}}  \\
    \bottomrule
    \end{tabular}
    }
    \label{fer_pp_fer}
\end{table}

\begin{figure}[!t]
    \centering
    \includegraphics[width=0.7\linewidth]{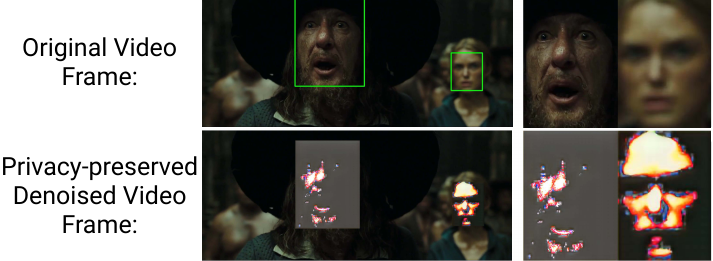}
    \caption{An example video frame from the DFEW dataset. 
    }
    \label{frame_example}
\end{figure}

\noindent\textbf{Privacy-preserved FER accuracy.} FER accuracies for each facial expression, using R(2+1)D and I3D across three datasets and the corresponding privacy-preserved results are reported in \cref{fer_pp_fer}.
\cref{main_result} summarizes the privacy-preserved FER performance of our approach (highlighted in gray) alongside the four baselines. Our method obtains the strongest FER accuracy.
Because~\cite{xu2024facial} requires closed-set identity labels, it is excluded from DFEW comparisons. For reference, we also report FER performance without privacy preservation (N-PP).
To assess stability, we run each method five times with different random seeds; the privacy–utility trade-offs are presented in \cref{five_ci}.
Confusion matrices for both FER and privacy-preserved FER appear in Supplementary Material~\cref{cm}, and qualitative examples of our outputs are shown in \cref{frame_example}. 
Overall, our method achieves the strongest privacy-preserved FER accuracy, particularly on open-set, in-the-wild datasets. Without privacy constraints, the two backbone models achieve comparable performance, and on RAVDESS the closed-set baselines offer only marginal advantages when using R(2+1)D.

\begin{figure}[!t]
    \centering
    \includegraphics[width=0.99\linewidth]{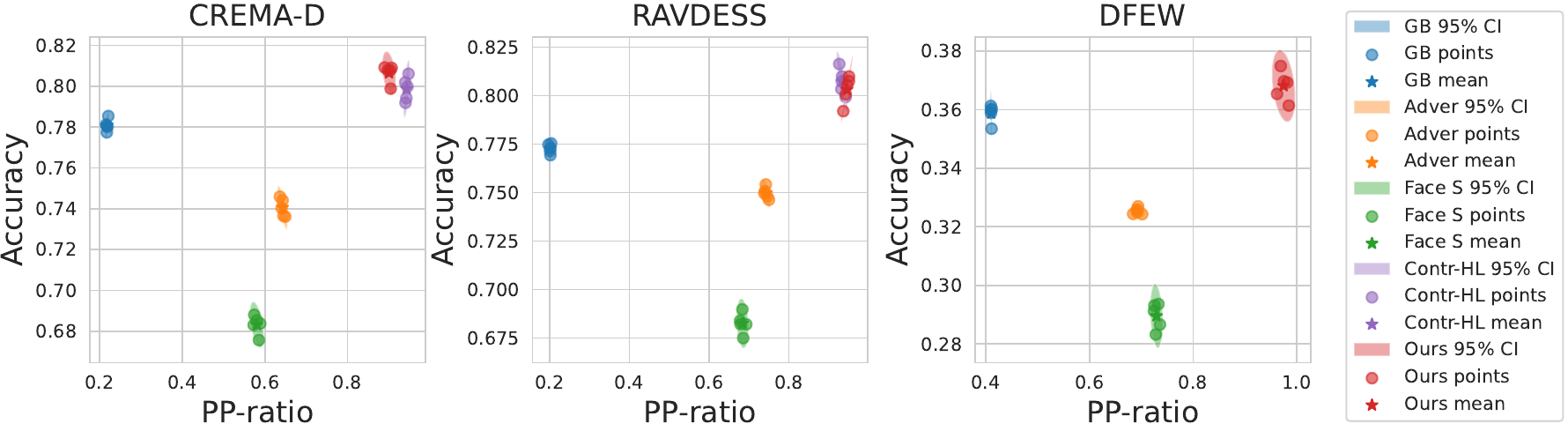}
    \caption{Privacy preservation ratio, $P_{pre}$ and FER accuracy tradeoff with 95\% confidence interval.}
    \label{five_ci}
\end{figure}

\begin{table}[!t]
    \centering
    \caption{Privacy-preserved and Denoised FER accuracy (\%). `N-PP' stands for \textit{no privacy preservation}, and it means standard FER task. In the GB row, the numbers in parentheses, (), represent the $\sigma$ hyperparameter values in Gaussian blurring. The `N/A' stands for Not Applicable.}
    \resizebox{\linewidth}{!}{
    \begin{tabular}{c|c|c|c|c|c|c}
    \toprule
       \multirow{2}{*}{Approach}  & \multicolumn{3}{c|}{Acc. of R(2+1)D $\uparrow$} & \multicolumn{3}{c}{Acc. of I3D $\uparrow$} \\ 
       & DFEW & CREMA-D & RAVDESS & DFEW & CREMA-D & RAVDESS \\ 
    \midrule
       N-PP & 43.14 & 89.15 & 86.44 & 42.43 & 88.26 & 87.73\\ 
    \midrule
       GB  & 35.89 (0.4) & 78.54 (0.5) & 77.54(0.6)& 36.32 (0.4)& 77.43 (0.5) &  78.21(0.6) \\ 
       Adver.~\cite{dave2022spact} & 32.44 & 74.59 & 75.08 &33.32 & 73.42 & 77.31\\ 
       Face S.~\cite{xu2022mobilefaceswap} & 29.31 & 67.56 & 68.39& 28.92 & 68.36 & 69.85\\ 
       Contr-HL~\cite{xu2024facial} & N/A &79.19 & \textbf{79.93} & N/A& 78.42 & 77.59\\ 
    \midrule
       \rowcolor{lightgray} Ours & \textbf{36.54} & \textbf{80.90} & 79.20 & \textbf{36.65} &\textbf{ 82.12} & \textbf{80.35}\\ 
    \bottomrule
    \end{tabular}
    }
    \label{main_result}
\end{table}

\noindent\textbf{Validation of privacy preservation.} 
In our proposed privacy preservation validation protocol, we first validate $f_{match}$ based on Rules 1-3 to verify its facial identity discrimination capability. Across the three datasets, 25,969 cases from DFEW, 4,464 from CREMA-D, and 2,343 from RAVDESS are generated under Rules 1-3. $f_{match}$ achieves accuracies of 0.9999, 1, and 1, respectively, confirming its reliability and consistency in extracting identity features.

\begin{table}[!t]
    \centering \label{pp_val}
    \caption{Privacy preservation ratio, $p_{pre}$. (): number of cases}
    \resizebox{0.9\linewidth}{!}{
    \begin{tabular}{c|c |c |c}
    \toprule
       Approach  & \makecell{DFEW {\footnotesize (82,746)}} & \makecell{CREMA-D  {\footnotesize (2466)}} & \makecell{RAVDESS {\footnotesize (1084)}}\\ \midrule

        GB & 0.4100 & 0.2206 & 0.2039 \\  
        Adver.~\cite{dave2022spact} & 0.6843 & 0.6363 & 0.7399 \\  
        Face S.~\cite{xu2022mobilefaceswap} & 0.7358 & 0.5856 & 0.6780 \\  
        Contr-HL~\cite{xu2024facial} & N/A & \textbf{0.9408} &\textbf{ 0.9419} \\  
        \midrule
        \rowcolor{lightgray} Ours &\textbf{0.9620} & 0.9071 & 0.9363 \\  
    \bottomrule
         
    \end{tabular}
    }
\end{table}

\begin{table}[!t]
    \centering
    \caption{SSIM and PSNR results.}
    \resizebox{\linewidth}{!}{
    \begin{tabular}{c|c|c|c|c|c|c}
    \toprule
       \multirow{2}{*}{Approach}  & \multicolumn{3}{c|}{SSIM $\downarrow$} & \multicolumn{3}{c}{PSNR $\downarrow$} \\ 
       & DFEW & CREMA-D & RAVDESS & DFEW & CREMA-D & RAVDESS \\ 
    \midrule
       GB & 0.9122& 0.9443& 0.9534&29.43 &32.85 & 33.91  \\ 

       Adver.~\cite{dave2022spact} &0.5120 & 0.5292 & 0.5145& 21.05 & 22.07& 21.09\\ 
       Face S.~\cite{xu2022mobilefaceswap} &0.4372 & 0.4934& 0.5002& 15.76& 16.31& 15.32\\ 
       Contr-HL~\cite{xu2024facial} & N/A& \textbf{0.4243}& \textbf{0.4398}& N/A& \textbf{11.49}& \textbf{12.09}\\ 
    \midrule
       \rowcolor{lightgray} Ours & \textbf{0.4076} & 0.4521&0.4432 &\textbf{11.47} &12.01 & 13.98\\ 
    \bottomrule
    \end{tabular}
    }
    \label{recoveryAttack}
\end{table}

Because~\cite{xu2024facial} requires a closed-set setting, it is excluded from DFEW comparisons. As shown in \cref{pp_val}, our method, leveraging both inter- and intra-video relationships, achieves strong privacy-preserving performance on in-the-wild datasets. Supplementary Material~\cref{pp_val_detailed} provides detailed case counts, numbers of correctly validated cases, and further analysis. Its performance on CREMA-D and RAVDESS is slightly lower than that of closed-set baselines, which is expected given that closed-set approaches retrain specifically on these datasets, whereas our model is trained only on DFEW and applied directly at inference. The in-the-wild dataset demonstrates stronger protection, likely because DFEW contains lower-quality facial regions, making identity recovery more difficult after privacy preservation.

Beyond our proposed validation ratio, we also compute SSIM and PSNR (reported in~\cref{recoveryAttack}). Our method surpasses all open-set approaches on these metrics but shows a small gap relative to closed-set methods. Note that SSIM and PSNR measure visual similarity rather than privacy, so they should be interpreted cautiously in this context.

\subsection{Ablation Study}
\label{ab}
\noindent\textbf{Different face tracking approaches.} 
The construction of knowledge priors and dataset generation relies on face-tracking outputs. Different tracking modules may affect overall performance. Specifically, we replace RetinaFace with MTCNN~\cite{zhang2016joint} and adopt SAMURAI~\cite{yang2024samurai} for tracking. 
As shown in~\cref{tracking}, under the same experimental setup described in~\cref{Implementation}, variations in the tracking component have only a marginal impact on downstream FER accuracy and privacy preservation.

\begin{table}[t]
    \centering
    \caption{The impact of various tracking approaches on the framework.}
    \resizebox{0.9\linewidth}{!}{
    \begin{tabular}{c|cc|c|c|>{\columncolor{gray!20}}c}
    \toprule
    
     \multicolumn{3}{c|}{Tracking Approach: }    & MTCNN+ArcFace & SAMURAI &  Ours\\ \midrule
     
      \multirow{4}{*}{DFEW}   & \multicolumn{2}{c|}{Num of Identity}& 20347& 26032& 25969 \\ \cline{2-6}
      & \multirow{2}{*}{\makecell{Acc. \\ (\%)}} & R(2+1)D& 36.42 & 36.71 & 36.54   \\ \cline{3-3}
      & & I3D & 36.60 & 36.27 & 36.65\\ \cline{2-6}
      &\multicolumn{2}{c|}{$P_{pre}$ (\%)}& 96.02 & 96.31 & 96.20\\ \midrule
      
      \multirow{3}{*}{CREMA-D}   & \multirow{2}{*}{\makecell{Acc. \\ (\%)}} & R(2+1)D& 80.85& 81.21 & 80.90\\ \cline{3-3}
      & & I3D & 81.97 & 82.49 &82.12\\ \cline{2-6}
      &\multicolumn{2}{c|}{$P_{pre}$ (\%)}& 90.98 & 91.03 &90.71\\ \midrule

      \multirow{3}{*}{RAVDESS}   & \multirow{2}{*}{\makecell{Acc. \\ (\%)}} & R(2+1)D& 79.93& 79.49&79.20\\ \cline{3-3}
      & & I3D & 81.04 & 80.89 & 80.35\\ \cline{2-6}
      &\multicolumn{2}{c|}{$P_{pre}$ (\%)}& 93.04 & 92.96 &93.63\\ \bottomrule

    \end{tabular}
    }
    
    \label{tracking}
\end{table}

\noindent\textbf{Different expression label of a batch.} 
In our framework, the knowledge priors and dataset generation stage enforces that each training batch contains a single expression label for privacy-preserving learning. However, if the second and subsequent triplets are replaced with different expressions (still sampled from different videos), the anchor identity remains unchanged. 
This approach satisfies both objective functions and can be used for training.
Therefore, in this ablation, from the second triplet, we select the anchor, positive, and negative samples using the same strategy but without conditioning on facial expression labels. The results are shown in~\cref{ab_label}.

\begin{table}[!t]
\centering
\caption{
(a) Ablation ignoring expression labels during dataset generation; only FER accuracy and $P_{pre}$ are evaluated. (b) FER accuracy and $P_{pre}$ on DFEW for varying $\alpha$.
}
\begin{subtable}[b]{0.49\linewidth}
\centering
\scriptsize
\resizebox{0.99\linewidth}{!}{\begin{tabular}{l|c|c|c}
\toprule
 & DFEW & CREMA-D & RAVDESS \\
\midrule
R(2+1)D Acc. & 33.98 & 79.02 & 77.64 \\
I3D Acc.     & 34.65 & 79.21 & 79.38 \\ \midrule
$P_{pre}$     & 95.98 & 91.29 & 93.98 \\
\bottomrule
\end{tabular}}\caption{} \label{ab_label}
\end{subtable}\hfill
\begin{subtable}[b]{0.49\linewidth}
\centering
\scriptsize
\resizebox{0.99\linewidth}{!}{\begin{tabular}{c|c|c|c}
\toprule
$\alpha$ & R(2+1)D Acc. & I3D Acc. & $P_{pre}$ \\
\midrule
0.5   & 0.2131 & 0.2034 & 0.9709 \\
0.2   & 0.2321 & 0.2398 & 0.9694 \\
0.1   & 0.2948 & 0.3037 & 0.9691 \\
0.05  & 0.3395 & 0.3487 & 0.9668 \\
0.005 & 0.3681 & 0.3639 & 0.9610 \\
\bottomrule
\end{tabular}}\caption{}\label{alpha}
\end{subtable}
\end{table}

\noindent\textbf{The choice of $\alpha$.}
The weight $\alpha$ controls the balance between the triplet loss and the BCE loss during privacy-preservation training. Using equal weights caused the FER model to fail, yielding near-random accuracy, as the triplet loss treated identical expressions as negatives.
Reducing the triplet-loss weight improved learning, and iterative testing identified $\alpha = 0.01$ as the most effective setting.
As shown in \cref{alpha}, smaller $\alpha$ values increase FER accuracy but slightly weaken privacy preservation; performance varies more noticeably at $\alpha = 0.1$ and $0.5$, while stabilizing at $\alpha = 0.01$.

\begin{table}[ht]
    \centering
    \caption{
    FER accuracy (\%) and privacy preservation ratio (\%), $P_{pre}$ on video clips without $f_{denoise}$. Numbers in parentheses show results with $f_{denoise}$ for comparison.
    }
    \resizebox{0.85\linewidth}{!}{
    \begin{tabular}{c|c|c|c}
    \toprule
        & DFEW&  CREMA-D    & RAVDESS  \\ \midrule
        R(2+1)D Acc.  & 27.55 (\textit{33.98})&   71.49 \textit{(79.02)}   & 73.54 \textit{(77.64)}  \\
        I3D Acc. & 28.31 (\textit{34.65})&  72.43 \textit{(79.21)}   & 73.11 \textit{(79.38)}  \\ \midrule
        $P_{pre}$& 95.01 \textit{(95.98)}&  87.59 \textit{(91.29)} &  90.13 \textit{(93.98)}\\
        \bottomrule
    \end{tabular}
    }
    
    \label{ab_denoise}
\end{table}

\noindent\textbf{Contribution of $f_{denoise}$.}
The denoising module $f_{denoise}$ enhances FER performance after privacy preservation. When FER training is performed directly on privacy-preserved outputs without denoising, FER accuracy drops substantially and the privacy-preservation ratio decreases slightly~\cref{ab_denoise}. 
This shows that $f_{denoise}$ benefits both utility and privacy. Trained as in \cref{denoise_train} to emphasize expression rather than full-image reconstruction, $f_{denoise}$ systematically shifts identity embeddings away from their originals, as illustrated by t-SNE plots in Supplementary Material \cref{denoising_analysis} and \cref{opd}, a detailed discussion.

%% file: 5.concl.tex
\section{Conclusion and Limitation}

We present a framework that preserves privacy while maintaining FER performance by leveraging inter- and intra-video knowledge priors. An open-set validation method is introduced to enable effective comparison with other privacy-preserving approaches. Experiments show that our method outperforms all baselines in both privacy protection and FER accuracy.

Despite these advances, several limitations remain. First, we use R(2+1)D and I3D for video-based FER rather than more advanced models; future work should examine their impact on privacy-preserving performance. Second, while the framework may extend to other privacy-sensitive tasks, such as human pose estimation, this remains unexplored. Finally, although the denoising module improves both FER and privacy preservation, its theoretical foundations require further study.

%% file: 6.supp.tex
\clearpage
\setcounter{page}{1}
\maketitlesupplementary

\section{Relationship between the Number of Triplets and Batch Size}
\label{sec:num_triplet}

We select the batch size of 256, because it can includes 85 triplets (containing 255 ($85 \times 3$) images plus one. The other batch size selections under our approach can be calculated using the following equation: $log_{2}(3n+1)=m$, where $n \in \mathbb{N}$ is the number of triplets, and $m \in \mathbb{N}$ is the exponent of $2$. 
We need to obtain the pairs of $m$ and $n$ for the selection. When $m$ is an even number, $2^m\equiv 1 (mod \ \ 3)$, and similarly, when $m$ is an odd number, $2^m\equiv 2 (mod \ \ 3)$. 
Because of $3n+1\equiv 1 (mod \ \ 3)$, the $m$ must be an even number. Let $m=2k$ (where $k\geq 0$ is an integer), and then we have $3n+1=2^{2k} = 4^{k}$. Solving for $n$, we get $n=(4^k-1)/3$. \cref{tab:batch_size} lists the first five selections of batch size and number of triplets. The number of triplets, $n$, starts from $1$, as at least one triplet is required in the training.
\begin{table}[ht]
    \centering
    \begin{tabular}{c|c|c|c}
         \toprule
         $k$& $m$ & $n$& Batch Size   \\ \midrule
         1&2&1&4\\
         2&4&5&16\\
         3&6&21&64\\
         4&8&85&256\\
         5&10&341&1024\\
         \bottomrule
    \end{tabular}
    \caption{Available options of batch size and number of triplets (n).}
    \label{tab:batch_size}
\end{table}

\section{Detailed Experiment Settings}\label{detailed_settings}

\subsection{Datasets}\label{detailed_dataset}
The proposed framework is designed for open-set, in-the-wild datasets that lack identity labels. For privacy preservation training, we select the DFEW dataset~\cite{jiang2020dfew} that contains 16,372 video clips. DFEW also provides 11,697 facial clips, each featuring a single facial identity per video. These facial clips are cropped and aligned with corresponding facial expression labels.
Additionally, we use CREMA-D~\cite{cao2014crema} with 7,442 videos and RAVDESS~\cite{livingstone2012ryerson} with 4,904 videos, two closed-set datasets containing both facial expression and facial identity labels. These datasets are utilized for privacy preservation inference, video-based FER training, and privacy preservation validation. Both datasets focus on facial expression performances, with controlled backgrounds, green in CREMA-D and white in RAVDESS. Each video features a single actor, ensuring consistency in identity-based evaluations.
The denoising module is pre-trained on the RAF-DB~\cite{li2017reliable} dataset,  which contains 29,672 real-world images. The pre-training uses the 7 facial expression classes.

\subsection{Detailed Implementation}\label{detailed_implementation}
\noindent \paragraph{Knowledge prior and dataset generation.}
For generating the knowledge priors, we select RetinaFace~\cite{deng2019retinaface} for face detection and alignment and ArcFace~\cite{deng2019arcface} for facial identity embedding extraction, with the cosine similarity threshold of 0.7. A total of 25,969 faces were detected and cropped across all frames of the 16,372 videos in DFEW. These faces form the dataset for privacy preservation training.

\noindent \paragraph{Privacy preservation training.}
For GPU optimization of privacy-preserving training, the batch size is constrained by the triplet structure. It has limited choices due to the initial data and the triplet. 
In our experiments, the batch size is set to 1,024, which includes 341 triplets. In addition, we analyze and compute some available options listed in the~\cref{sec:num_triplet}. In this training phase, we employ U-Net as $f_{pp}$ for privacy preservation and ArcFace~\cite{deng2019arcface} as $f_{e}$ for facial feature extraction, following the procedure in~\cref{pp_train}. 
The loss weight $\alpha$ is set to 0.01, and the training epoch number is set to 400. The random seed is fixed at 42. Once trained, $f_{pp}$ is used to infer privacy-preserved versions of both the cropped faces from all 16,372 videos and the 11,697 facial clips for denoising, FER training, and falsification-based privacy validation. Inferred facial clips are used for denoising and FER training~(\cref{fig:framework}, Step 3), while inferred cropped faces are used for privacy preservation validation~(\cref{fig:framework}, Step 4).

\noindent\paragraph{Denoising Pre-training}
The denoising model, $f_{denoise}$, is pre-trained following \cref{denoise_train} on the RAF-DB dataset with its 7-class facial expression labels and is applied to the privacy-preserved facial clips before they are used for FER training.

\noindent\paragraph{Video-based FER training}
For video-based FER training, we select the ``set\_1'' train-test split provided by DFEW~\cite{jiang2020dfew}.
There are 7,442 and 4,904 faces in the CREMA-D and RAVDESS, respectively. These faces undergo privacy preservation using the $f_{pp}$ model trained on DFEW, with only their bounding boxes used for inference. We assign 30\% of the videos in each dataset as the test set. For the FER task, we use R(2+1)D~\cite{tran2018closer} and I3D~\cite{carreira2017quo}, both pre-trained on Kinetics-400~\cite{kay2017kinetics}. The data first pass through $f_{denoise}$ with its weight frozen, and then are fed into the FER model $f_{pp}$ (either R(2+1)D or I3D). By optimizing $f_{pp}$ with 7-class facial expression labels, it learns the classification capability.

\noindent\paragraph{Privacy Validation}
In privacy preservation validation, the recovery model $f_{reco}$ is a U-Net trained in two phases: one for $\langle i_{org}, i_{pp} \rangle$ pairs and another for $\langle i_{org}, i_{dpp} \rangle$ pairs. 
$G_{cases}$ implements all rules from~\cref{strategy}. After $f_{reco}$ training, it infers both $D_{pp}$ and $D_{dpp}$ datasets for $G_{cases}$ to generate the data for Rules 6 and 7. Before this inference, there are data for Rule 1 to 5. $G_{cases}$ utilizes the relationships between data from the same or different videos to generate the ground truth. After the falsification case generation, the process proceeds to matching stage. The matching function $f_{match}$ utilizes a pre-trained ArcFace~\cite{deng2019arcface} model for face recognition with a binary classification head: 0 for different identity and 1 for the same identity. We then compare the results with the ground truth, count the numbers of correct and incorrect cases, and derive the privacy-preservation ratio.

\subsection{Detailed Baseline Approaches}\label{dba}
We evaluate four baseline methods for privacy preservation. 
The first three, selected from Rows 2-3 in \cref{pp_compare}, are applicable in an open-set scenario. In our comparative experiments, these methods were not integrated into our framework's privacy preservation component. Instead, FER training was conducted after applying blurring and face swapping. The last baseline methods were implemented as described in their respective papers. We applied the denoising module and recovery attack from our framework to all baselines except~\cite{xu2024facial}, which already includes a ``feature compensator'', during evaluation.

\noindent\textbf{Blurring.}
We use the Gaussian blur method as a baseline for privacy preservation.
The function is defined as $GB(x,y)= \frac{1}{2\pi \sigma^2}exp({ - \frac{x^2+y^2}{2\sigma^2} })$, where $(x,y)$ are pixel coordinates relative to the filter center, and $\sigma$ controls the blur intensity. By adjusting $\sigma$ , we achieve a comparable FER performance to our approach, denoted as ``GB''. The chosen $\sigma$ values for each dataset are provided in \cref{main_result}.

\noindent\textbf{Adversarial privacy-preserved approach.}
We implement an adversarial privacy-preserved approach from~\cite{dave2022spact}, which does not require privacy attribute labels and is suitable for open-set environments. In the results, this approach is referred to as ``Adver.''.

\noindent\textbf{Face swapping.}
Face swapping replaces an individual's face with a synthetic identity to obscure the original identity. We select the state-of-the-art MobileFaceSwap method~\cite{xu2022mobilefaceswap} to replace all faces in the dataset with a fake identity. This approach is referred to as ``Face S.'' for short.

\noindent\textbf{Controlled high- and low-frequency approach.}
The closed-set privacy preservation approach from~\cite{xu2024facial} utilizes controllers trained with privacy attribute labels. Since this method requires labeled privacy attributes, it is only applicable to the CREMA-D and RAVDESS datasets. In the result tables, it is referred to as ``Contr-HL''.

\section{Additional Experimental Results}
\subsection{Original, Privacy-preserved and Denoised Identity Relationship}\label{opd}
\begin{figure}[t]
    \centering
    \includegraphics[width=0.8\linewidth]{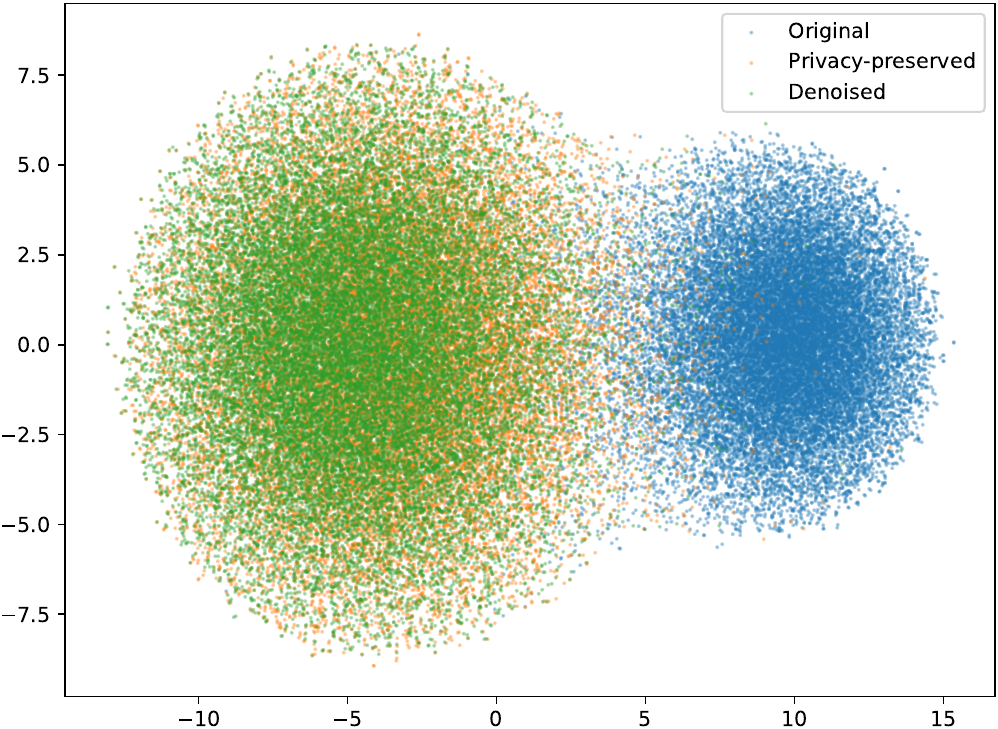}
    \caption{t-SNE of the original, privacy-preserved and denoised privacy-preserved identity embedding from DFEW. It indicates that the model systematically shifts identity features away from their original identities.}
    \label{denoising_analysis}
\end{figure}

We extracted identity embeddings from DFEW at three stages, including the original, the privacy-preserved and the denoised privacy-preserved, and computed a t-SNE projection. As visualized in \cref{denoising_analysis}, the results indicate that the model modifies the identity features of the images along a specific direction, shifting them further away from the original identities.

As shown in the \cref{denoising_analysis}, the identity embeddings at different stages form three distinct distributions, while a small portion of samples appear overlapped across stages. However, such visual overlap does not necessarily indicate similarity at the sample level, as t-SNE is designed for qualitative visualization and does not preserve pairwise distances in the original embedding space. To more accurately quantify identity changes at the individual level, we therefore introduce a sample-wise metric, termed Individual Embedding Displacement.

Given a sample $i$, let $z_i^o, z_i^p, z_i^d \in \mathrm{R}^{512}$ denote its identity embedding at three stages of original, privacy-preserved and denoised privacy-preserved. Since embeddings are sample-wise aligned across stages, we define the Individual Embedding Displacement (IED) directly measuring identity variation at the individual level in the original high-dimensional embedding space between two random stages ($m,n$) of these three stages:
\begin{equation}
    IED^{m,n}_i = || z_i^m - z_i^N ||_2
\end{equation}
\begin{figure}
    \centering
    \includegraphics[width=0.9\linewidth]{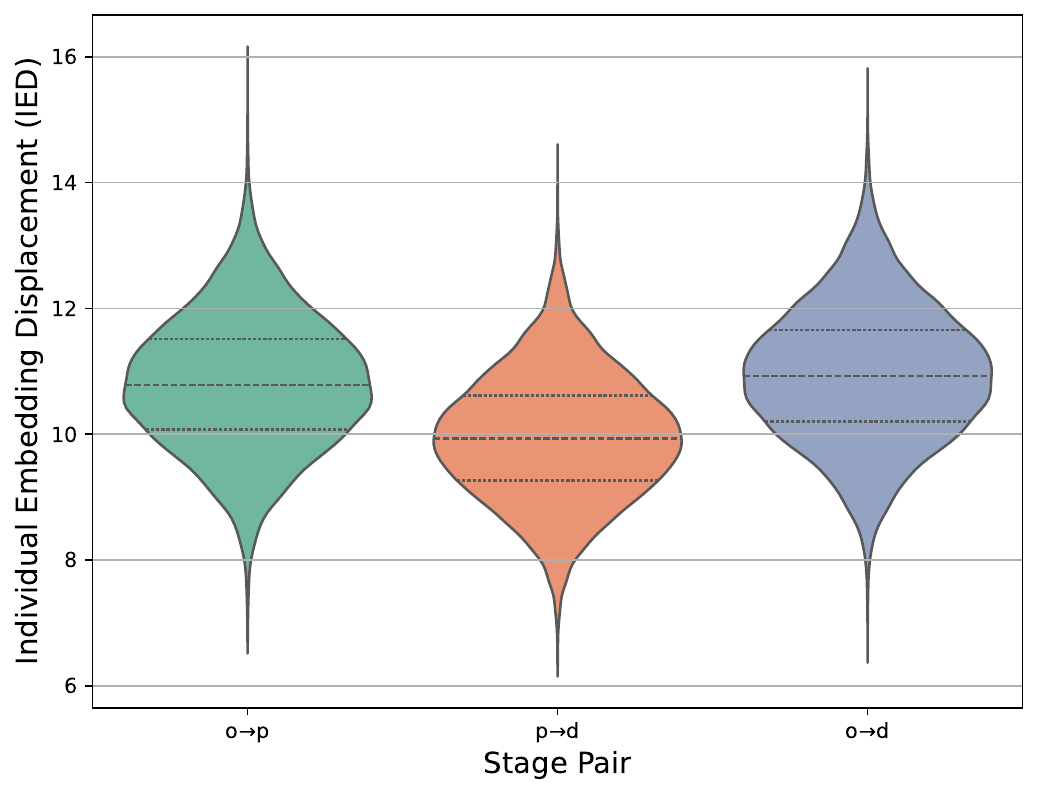}
    \caption{Distribution of IED Across Stages}
    \label{ied}
\end{figure}

\cref{ied} shows the distribution of IED across different identity embedding stages. Even for samples that appear overlapped in t-SNE, the IED values indicate substantial movement in the original high-dimensional embedding space.

\subsection{Detailed Falsification Cases}
\begin{table}[!t]
\centering
\caption{\small Statistics of Rule 4–7 cases, including correct validations and privacy-preservation (PP) ratios.}
\resizebox{0.9\linewidth}{!}{
    \begin{tabular}{c|c|cccc|c}
    \toprule
    \multicolumn{2}{c|}{Dataset} & Rule 4 & Rule 5 & Rule 6 & Rule 7 &  PP Ratio \\ \midrule
    \multirow{5}{*}{DFEW} 
     & \# Cases   & 25969 & 25969 & 25969 & 25969 &  \\
     & GB      & 15842 & 16049 &  5291 &  5411 & 0.4100 \\
     & Adver.  & 21437 & 21625 & 13975 & 14046 & 0.6843 \\
     & Face S  & 22837 & 22981 & 14675 & 14846 & 0.7253 \\ 
\rowcolor{lightgray}     &  Ours    & 24897 & 25297 & 24681 & 25052 & \textbf{0.9620} \\ \midrule
    \multirow{6}{*}{CREMA-D} 
     & \# Cases    & 4464 & 4464 & 4464 & 4464 &  \\
     & GB       & 1681 & 1706 &  436 &  116 & 0.2206 \\
     & Adver.   & 3409 & 3485 & 2189 & 2279 & 0.6363 \\
     & Face S   & 3087 & 3164 & 2080 & 2126 & 0.5856 \\
     & Contr-HL & 4408 & 4494 & 3910 & 3987 & \textbf{0.9408} \\
  \rowcolor{lightgray}  &   Ours     & 4108 & 4127 & 3952 & 4010 & 0.9071 \\ \midrule
    \multirow{6}{*}{RAVDESS} 
     & \# Cases    & 2343 & 2343 & 2343 & 2343 &  \\
     & GB       &  789 &  698 &  159 &  265 & 0.2039 \\
     & Adver.   & 1806 & 1902 & 1567 & 1659 & 0.7399 \\
     & Face S   & 1645 & 1697 & 1471 & 1538 & 0.6776 \\
     & Contr-HL & 2256 & 2380 & 2068 & 2124 & \textbf{0.9420} \\ 
\rowcolor{lightgray}     &  Ours     & 2201 & 2299 & 2101 & 2174 & 0.9363 \\ \bottomrule
    \end{tabular}
}

\vspace{-1em}
\label{pp_val_detailed}
\end{table}
\cref{pp_val_detailed} reports the number of cases generated under Rules 4–7, the number of correctly validated cases (i.e., predictions aligned with ground truth), and the resulting validation ratios for all privacy-preservation methods.
\section{Confusion Matrix}\label{cm}
\subsection{FER Classification without Privacy Preservation} 

\begin{figure}[ht]
    \centering
    \includegraphics[width=0.9\linewidth]{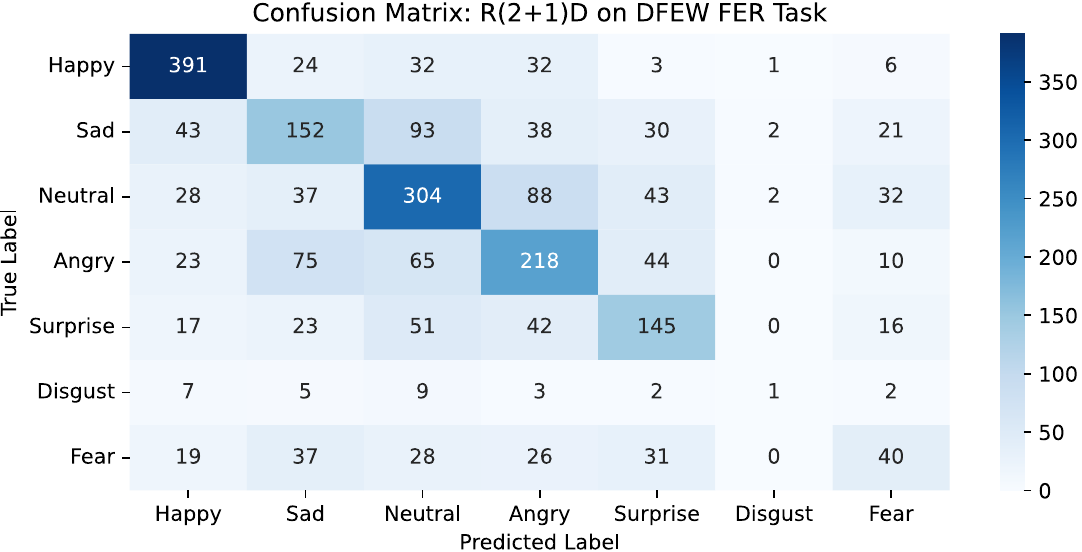}
    \caption{FFE task on DFEW dataset using R(2+1)D model \textit{without privacy preservation}}
    \label{fer-dfew-r3D}
\end{figure}

\begin{figure}[ht]
    \centering
    \includegraphics[width=0.9\linewidth]{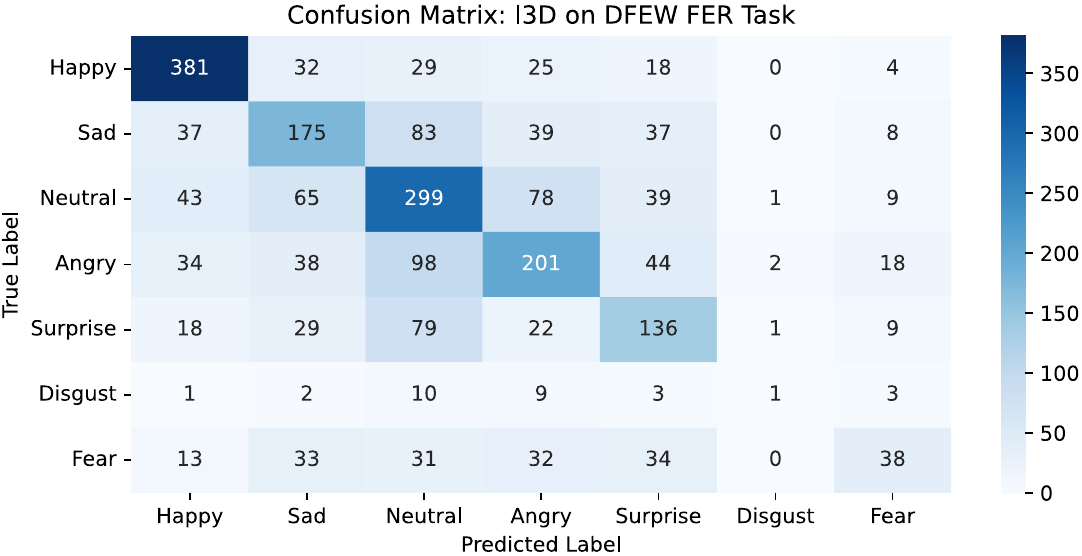}
    \caption{FFE task on DFEW dataset using I3D model \textit{without privacy preservation}}
    \label{fer-dfew-i3D}
\end{figure}

\begin{figure}[ht]
    \centering
    \includegraphics[width=0.9\linewidth]{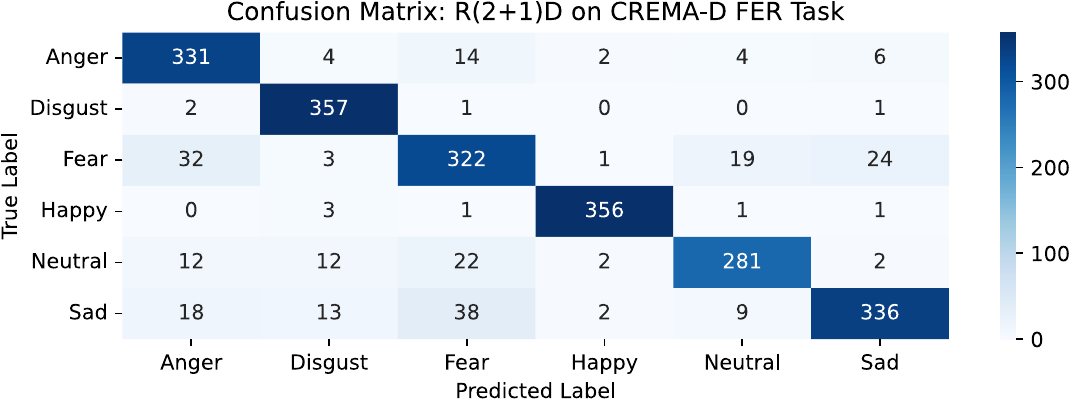}
    \caption{FFE task on CREMA-D dataset using R(2+1)D model \textit{without privacy preservation}}
    \label{fer-cremad-r3d}
\end{figure}

\begin{figure}[ht]
    \centering
    \includegraphics[width=0.9\linewidth]{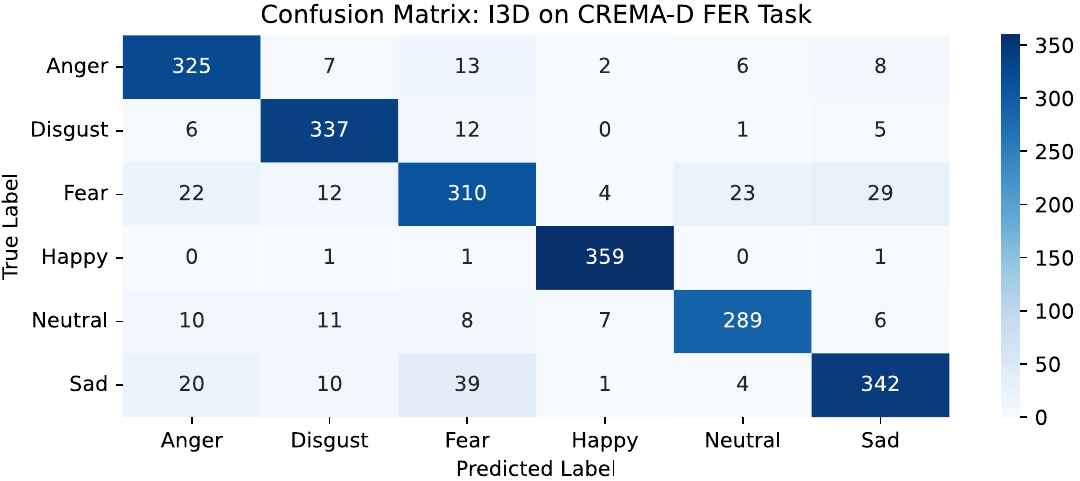}
    \caption{FFE task on CREMA-D dataset using I3D model \textit{without privacy preservation}}
    \label{fer-cremad-i3d}
\end{figure}

\begin{figure}[ht]
    \centering
    \includegraphics[width=0.9\linewidth]{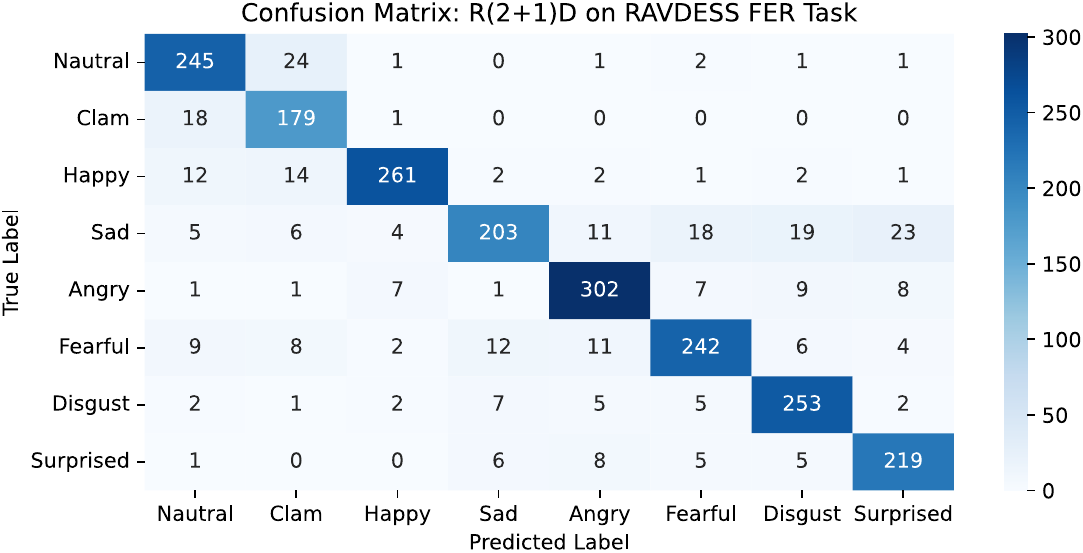}
    \caption{FFE task on RAVDESS dataset using R(2+1)D model \textit{without privacy preservation}}
    \label{fer-ravdess-r3d}
\end{figure}

\begin{figure}[ht]
    \centering
    \includegraphics[width=0.9\linewidth]{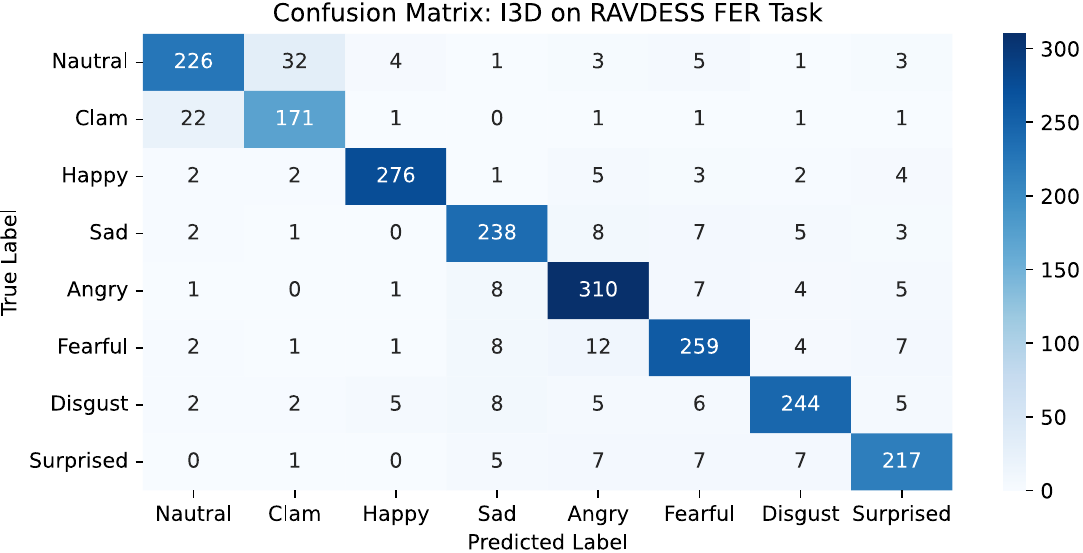}
    \caption{FFE task on RAVDESS dataset using I3D model \textit{without privacy preservation}}
    \label{fer-ravdess-i3d}
\end{figure}

\cref{fer-dfew-r3D,fer-dfew-i3D,fer-cremad-r3d,fer-cremad-i3d,fer-ravdess-r3d,fer-ravdess-i3d} provide the confusion matrix of R(2+1)D and I3D models on the test dataset on DFEW, CREMA-D and RAVDESS.

\subsection{Privacy Preserved FER Classification}

\begin{figure}
    \centering
    \includegraphics[width=0.9\linewidth]{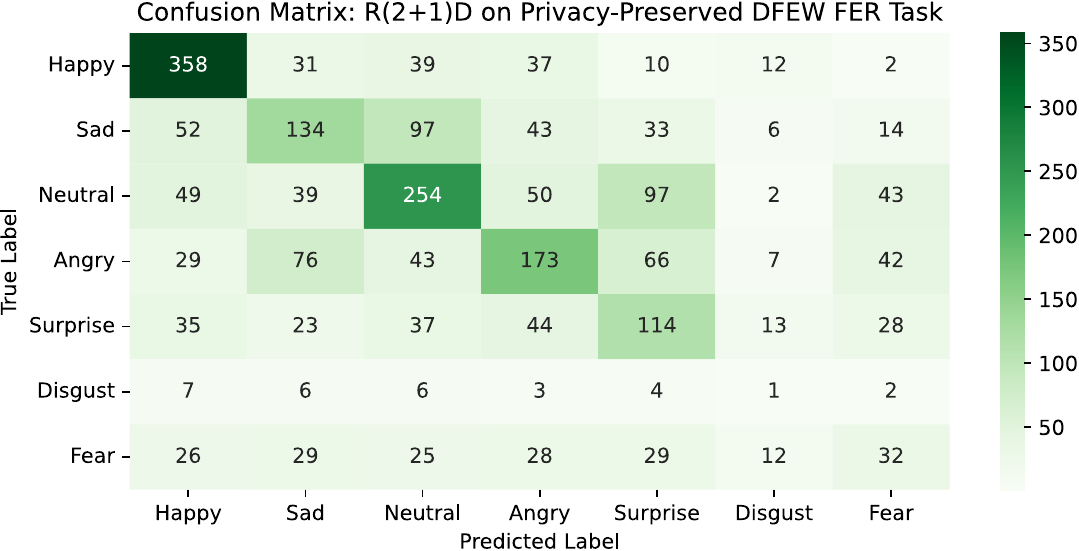}
    \caption{FFE task on \textit{privacy-preserved and denoised} DFEW dataset using R(2+1)D model}
    \label{pp-dfew-r3d}
\end{figure}

\begin{figure}
    \centering
    \includegraphics[width=0.9\linewidth]{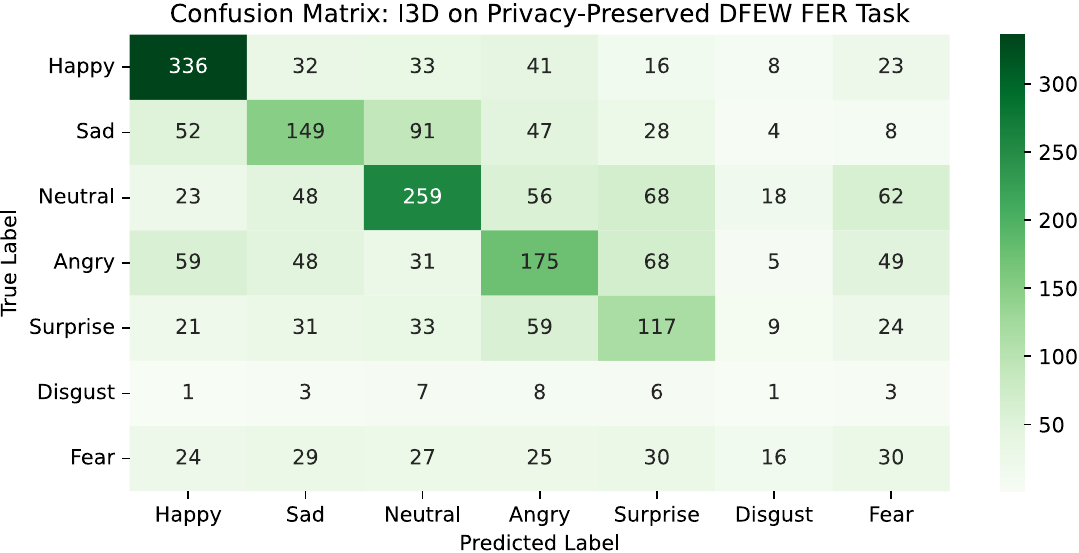}
    \caption{FFE task on \textit{privacy-preserved and denoised} DFEW dataset using I3D model}
    \label{pp-dfew-i3d}
\end{figure}

\begin{figure}
    \centering
    \includegraphics[width=0.9\linewidth]{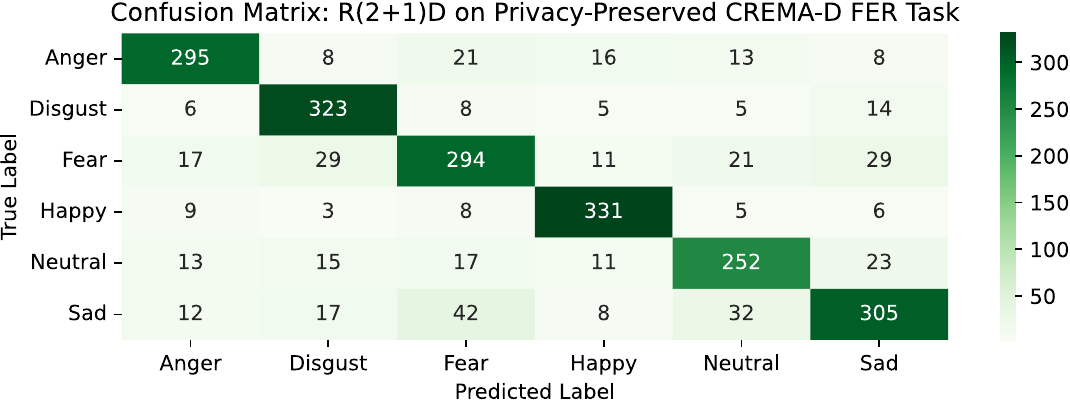}
    \caption{FFE task on \textit{privacy-preserved and denoised} CREMA-D dataset using R(2+1)D model}
    \label{pp-cremad-r3d}
\end{figure}

\begin{figure}
    \centering
    \includegraphics[width=0.9\linewidth]{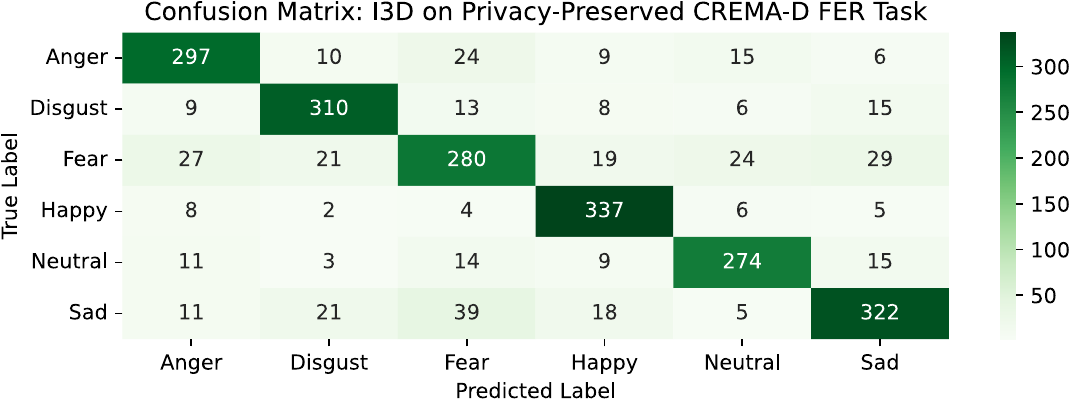}
    \caption{FFE task on \textit{privacy-preserved and denoised} CREMA-D dataset using I3D model}
    \label{pp-cremad-i3d}
\end{figure}

\begin{figure}
    \centering
    \includegraphics[width=0.9\linewidth]{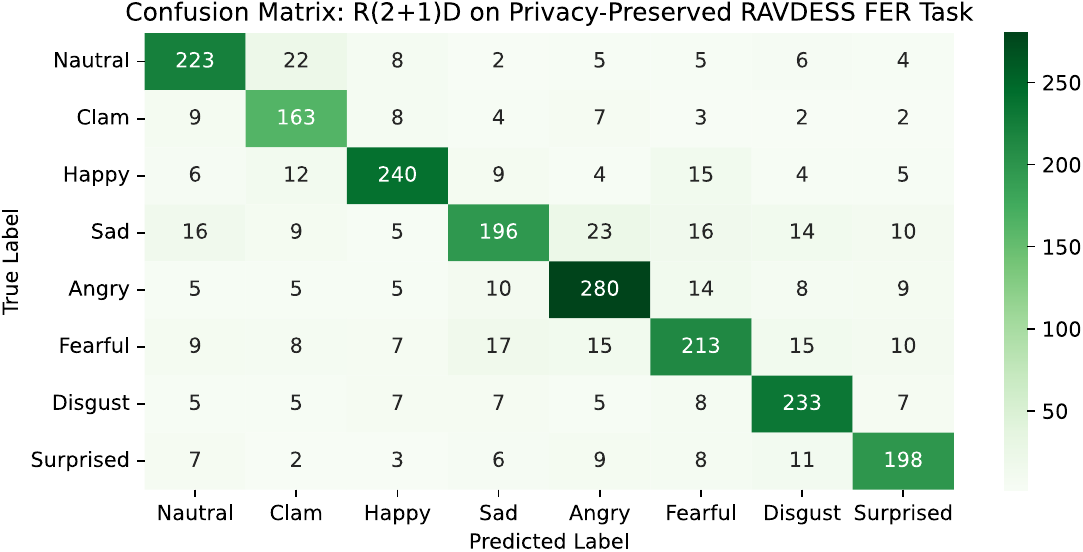}
    \caption{FFE task on \textit{privacy-preserved and denoised} RAVDESS dataset using R(2+1)D model}
    \label{pp-RAVDESS-r3d}
\end{figure}

\begin{figure}
    \centering
    \includegraphics[width=0.9\linewidth]{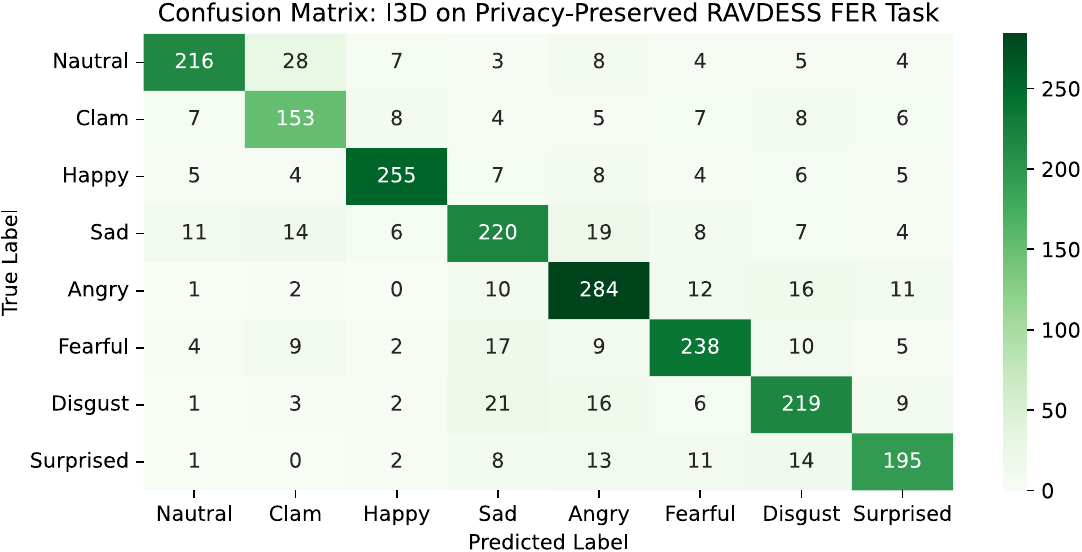}
    \caption{FFE task on \textit{privacy-preserved and denoised} RAVDESS dataset using I3D model}
    \label{pp-RAVDESS-i3d}
\end{figure}

\cref{pp-dfew-r3d,pp-dfew-i3d,pp-cremad-r3d,pp-cremad-i3d,pp-RAVDESS-r3d,pp-RAVDESS-i3d} provide the confusion matrix of R(2+1)D and I3D models on the test dataset on privacy-preserved DFEW, CREMA-D and RAVDESS.

